\documentclass{article}

\usepackage[preprint]{nips_2018}
\usepackage{caption} 
\usepackage{float}
\usepackage{amsthm}
\usepackage{xcolor}
\usepackage{booktabs}
\usepackage[utf8]{inputenc} 
\usepackage[T1]{fontenc}    
\usepackage{hyperref}       
\usepackage{url}            
\usepackage{booktabs}       
\usepackage{amsfonts}       
\usepackage{nicefrac}       
\usepackage{microtype}      
\usepackage{graphicx}
\usepackage{amsmath,amssymb}
\usepackage{amsfonts}       
\usepackage{color}
\usepackage{transparent}
\usepackage{booktabs}
\usepackage{natbib}

\usepackage{amsmath}
\usepackage{enumerate}

\usepackage{nicefrac}       
\usepackage{amsmath}
\DeclareMathOperator*{\argmax}{arg\,max}
\newcommand{\ra}[1]{\renewcommand{\arraystretch}{#1}}
\usepackage{algorithm2e}
\usepackage{booktabs}
\usepackage{siunitx}
\newtheorem{mydef}{Definition}

\usepackage{placeins}

\title{Bayesian Optimisation with Gaussian Processes \\ for Premise Selection}

\author{
  Agnieszka Słowik\\
  University of Cambridge\\
 \\
  \And
  Chaitanya Mangla\\
  University of Cambridge\\
    \And
  Mateja Jamnik\\
  University of Cambridge\\
     \And
  Sean B. Holden\\
  University of Cambridge\\
    \And
  Lawrence C. Paulson\\
  University of Cambridge\\
   }

\begin{document}

\maketitle

\begin{abstract}
Heuristics in theorem provers are often parameterised. Modern theorem provers such as Vampire \citep{Kovacs:2013:FTP:2958031.2958033} utilise a wide array of heuristics to control the search space explosion, thereby requiring optimisation of a large set of parameters. An exhaustive search in this multi-dimensional parameter space is intractable in most cases, yet the performance of the provers is highly dependent on the parameter assignment. In this work, we introduce a principled probablistic framework for heuristics optimisation in theorem provers. We present results using a heuristic for premise selection and The Archive of Formal Proofs (AFP) \citep{Jaskelioff-Merz-AFP05} as a case study.

\end{abstract}{} 

\section{Introduction}

Theorem provers use heuristics at various points in their operation, such as in search control and premise selection. These heuristics often have parameters that greatly influence the practical performance of a prover. Existing approaches to selecting such parameters require human supervision, rules of thumb or extensive testing. Such testing is often conducted on large theory sets, and is thus computationally expensive. Every assignment to the collection of parameters forms a point in parameter space, and as the parameters grow in number and range, an exhaustive search for optimal parameters becomes infeasible. An alternative is to sparsely navigate the space of parameters in search of the optimal point. We argue that probabilistic search enables efficient and automated optimisation of parameterised heuristics in theorem proving.

A simple probabilistic approach to parameter selection is to use a variation of \textit{$\epsilon$-greedy search} \citep{Sutton1998}. Given a metric that determines the value of each parameter combination within the parameter range, the agent starts at a random point and greedily searches the local neighbourhood. The best point found becomes the starting point for the next iteration. As a result, the agent is approaching a local optimum. However, with probability equal to $\epsilon$, the agent randomly draws a new point for exploration. In theory, the $\epsilon$-greedy search is an exhaustive search given unlimited time, and therefore in the limit it should give us the global optimum. In practice, it makes inefficient use of resources since the points it tests tend to be densely clustered. The key issue here is that the knowledge gained by testing any point is discarded.

In heuristics used in theorem proving, for instance premise selection algorithms, the objective function that maps parameter assignment to the number of proofs found is not given explicitly. What we care about is to find the parameter assignment that maximises the practical performance of a prover, and this should be done in a minimum number of evaluations to avoid the costs associated with exhaustive testing. The knowledge gained from every tested point in the parameter space can be used to reduce the uncertainty about the shape of the objective function. We can use our
understanding of the objective function within this uncertainty to make an informed decision about the next point to test. Every test further reduces our uncertainty. Furthermore, we often have a priori knowledge of certain aspects of the objective function. For example, we may have an idea of how close we expect neighbouring points to be. If we construct a probabilistic model of the objective function with this a priori knowledge, we can then use the knowledge gained from testing arbitrary points to increase our certainty, thereby improving our prediction of the next best point to test. \textit{Bayesian Optimisation} \citep{10.1007/3-540-07165-2_55} is a principled method for this purpose and \textit{Gaussian Processes} \citep{Rasmussen:2005:GPM:1162254} provide a means for probabilistic modelling of functions using prior knowledge and machine learning.

Here, we explore Bayesian Optimisation with Gaussian Processes as a general solution to efficient heuristics tuning in automated theorem proving. We conduct a case study in premise selection using a state-of-the-art heuristic Sumo Inference Engine \citep{10.1007/978-3-642-22438-6_23}. Our framework based on Gaussian Processes took up to nine minutes to find the optimal set of parameters in ten AFP articles. The premises recommended by the optimised SInE were sufficient to prove 85.3\% of the conjectures using Sledgehammer \citep{Bohme:2010:SJD:2176669.2176681}. 

\section{Premise Selection}
In the experiments presented in this extended abstract we focused on the task of \textit{premise selection} which can be defined as follows:

\begin{mydef}
Given a set of premises $\mathbb{P}$, an ATP system $\mathcal{A}$ and a new conjecture $C$, select the premises from $\mathbb{P}$ that will most likely lead to a proof of $C$ by $\mathcal{A}$.
\end{mydef}

The size of the modern mathematical corpora creates a ``needle in a haystack'' problem for automated theorem provers. Only a small subset of available premises is relevant to any given conjecture. For instance, Open CYC \citep{matuszek_introduction_2006} contains over 3 million axioms while each of the problems has a proof involving up to 20 premises. 

\emph{Sumo Inference Engine (SInE)} is a simple heuristic-based premise selection algorithm introduced to optimise reasoning in large theories. The algorithm aims to estimate the importance of function and predicate symbols based on their frequencies in the conjecture and in the premises at hand. The least frequent symbols indicate a \textit{trigger relation} between the goal statement and a premise. This basic variant of the trigger relation is defined as follows:

Let $occ(s)$ be the number of premises in which the symbol $s$ occurs and $S$ the set of all symbols in a premise $p$. We define the least general symbol $s'$ as the symbol for which:

\begin{equation}
    \forall s \in S: occ(s') \leq occ(s) 
\end{equation}

and use it as a trigger for $p$. If the symbol $s'$ appears in the conjecture, the algorithm selects the premise $p$. This basic heuristic suffers from low robustness since small changes in the number of frequencies can lead to a loss of important premises. The heuristic was thus extended in the following way:

\begin{equation}
    trigger(s') \iff \{occ(s') \le g\} \vee \{\forall s \in S: occ(s') \le t \cdot occ(s) \}
\end{equation}

\noindent where the parameters $t \geq 1$, $g \geq 1$ are referred to as the \textit{tolerance threshold} and \textit{generality threshold}, respectively. 
\newline
\newline
Finally, premises triggered by the goal may contain symbols that lead to other relevant premises. This introduces another parameter $k \geq 0$ referred to as \textit{depth}. It leads to the inductive construction of triggering symbols and premises:
\begin{enumerate}
    \item All  symbols $s$ of the goal are 0-step triggered.
    \item If $s$ is $k$-step triggered and it triggers a premise $p$, then $p$ is $k+1$-step triggered.
    \item If $p$ is $k$-step triggered and $s$ occurs in $p$, then $s$ is $k$-step triggered.
\end{enumerate}

The algorithm is therefore parameterised by one continuous parameter $t$ and two discrete parameters $g$ and $k$. As it was demonstrated in \citep{10.1007/978-3-642-22438-6_23}, those parameters greatly influence the performance of the algorithm. The disparities between the optimal parameter assignment depending on the problem set can be significant. For instance, premise selection on SUMO and CYC problems from the TPTP library benefits from setting the depth parameter $k$ to infinity, while using a problem set from Mizar results in selection of the sufficient number of premises with the depth limit set to 1. Bayesian optimisation framework provides an automatic suggestion of the optimal parameter assignment based on the problem set.

\section{Bayesian Optimisation}

Bayesian optimisation methods construct a probabilistic model of the objective function $f$ and use it to determine informative sample locations. Under some prior on $f$, the points in the parameters space are repeatedly evaluated based on the posterior mean and variance predictions.

In our setting, the unknown objective function represents the usefulness of SInE given a parameter assignment. We assume this function was sampled from a Gaussian process. As we evaluate the performance of SInE given the point in the parameter space, the Bayesian Optimisation framework improves the posterior distribution for the objective function as the agent becomes more certain of which regions are worth exploring. In our implementation we choose the point in the parameter space to be evaluated in the next iteration based on the posterior distribution and upper confidence bound of a Gaussian process which is one of the standard methods referred to as the \emph{Gaussian Process-Upper Confidence Bound (GP-UCB)} algorithm \citep{Srinivas:2010:GPO:3104322.3104451}. The subsections below describe this approach in more detail.

\subsection{Problem statement}
We want to maximise an unknown objective function $f: D \rightarrow \mathbb{R}$ which is correlated to the number of conjectures an ATP system would be able to prove given the premises selected by SInE. At each iteration $i$, we choose a point $x_{i} \in D$ and evaluate the current estimate of the objective function. The goal is to find the solution to the following expression (as rapidly as possible):
\begin{equation}
x^\star=\argmax_{x \in D}{f(x)} .
\end{equation}
\subsection{Gaussian Processes}
A Gaussian Process is a distribution over functions specified by a mean $\mu$ and a covariance function $\kappa$:
\begin{equation}
    f(x) \sim GP (\mu(x), \kappa(x, x')).
\end{equation}
Common choices of covariance function include the finite dimensional linear, squared exponential and Mat\'ern kernels \citep{phdthesis}. We used the Mat\'ern kernel which can be seen as a generalization of the Gaussian radial basis function. 

\subsection{Gaussian Process-Upper Confidence Bound}
\label{GPUCB}
We define the \emph{upper confidence bound} in the maximisation problem as:
\begin{equation}
    UCB(x) = \mu(x) + \beta\sigma(x)
\end{equation}
where $\beta \geq 0$.
The first part of the update rule favours the points $x$ which are likely to give a high reward in terms of the objective function. The second part prefers the points where the function $f$ is uncertain, thereby negotiating the trade-off between exploitation and exploration. The amount of exploration is controlled by the constant $\beta$.

\section{Experiments}

\subsection{Dataset}
The Archive of Formal Proofs (AFP) \citep{Jaskelioff-Merz-AFP05} is a rapidly expanding\footnote{See the latest AFP statistics at \url{www.isa-afp.org/statistics.shtml}} collection of proofs formalised in Isabelle \citep{Nipkow:2002:IPA:1791547}. The repository is organized in the manner of a scientific journal. We used a parsed version of the dataset that meets the input requirements of MaSh \citep{10.1007/978-3-642-39634-2_6}, the machine learning premise selector currently implemented in Isabelle. Here, we report the results on 10 articles containing various theories and of sizes ranging from around 100 to around 1500 conjectures. 

Each conjecture was paired with a history of premises extracted from Sledgehammer logs \citep{Bohme:2010:SJD:2176669.2176681} that were used to determine which lemmas are needed to prove a goal.

\subsection{Evaluation metrics}

We have approximated the impact of premise selection on a theorem prover by using two evaluation metrics that are computed using SInE recommendations and the correct set of premises. The correct set of premises was obtained from Sledgehammer.

In premise selection, it is acceptable to provide more premises than necessary to prove a conjecture in order to minimise the risk of missing a key lemma. However, the main purpose of filtering is to lower the cost of considering irrelevant lemmas, and so an efficient algorithm should minimise the number of unnecessary recommendations. 

To let this trade-off guide the optimisation process, we propose a metric based on \textit{precision} and \textit{recall}. In our setting, recall is represented by the ratio of the relevant premises recommended by the selection algorithm to the total number of premises needed for the proof. We experimented with several approaches to expressing precision in our setting and found that using a ratio of relevant premises recommended by SInE to an expression that increases exponentially with the size of the recommended set:
\begin{enumerate}
    \item leads to an objective function that is strongly correlated with the main goal of recommending all of the premises needed to prove a conjecture;
    \item favours a small number of redundant recommendations over the risk of missing one of the key premises, but penalises large sets of recommendations;
    \item reduces sparsity that arises from directly optimising the number of theorems proved, which improves the optimisation process.
\end{enumerate}

This metric is computed individually for each conjecture as follows:

Assuming we have a set of $n$ conjectures, let $i = 1, 2, \dots, n$ be the index of a conjecture to be proved. Let $P_{i}$ be a non-empty set of lemmas required to prove the conjecture $i$ and $\tilde{P_{i}}$ a set of lemmas recommended by a premise selection algorithm. Here, $P_{i}$ is the set of premises used by Sledgehammer to prove the conjecture $i$, and $\tilde{P_{i}}$ refers to the set of premises recommended by SInE. If $|\tilde{P_{i}}|\neq 0$:

\begin{equation}
    S_{i} = \frac{|\tilde{P_{i}} \cap P_{i}|}{|P_{i}|} + \frac{|\tilde{P_{i}} \cap P_{i}|}{2^{|\tilde{P_{i}}|}}.
\end{equation}

For $|\tilde{P_{i}}| = 0$ we set the score $S_{i}$ to zero since the conjecture could not be proved. After computing the value of $S_{i}$ for each conjecture, we define the objective function across the whole dataset as follows:
\begin{equation}
    S = \sum_{i=1}^{n} S_{i}.
\end{equation}

At the testing stage we evaluate the algorithm based on the number of conjectures that would be proved in practice by Sledgehammer using the premises recommended by SInE. We assume that all of the premises used by Sledgehammer are necessary to prove the conjecture whereas in practice the prover might be able to find an alternative solution that requires a different set of premises. Consequently, this testing metric will tend to underestimate the number of conjectures proved using the SInE recommendations. 

\subsection{Results}
\subsubsection{Grid Search and Bayesian Optimisation}
Gaussian Processes assume continuous input variables. Discrete-valued parameters require additional approximations. A common approach is to use a surrogate continuous variable for each discrete parameter, and round its value to the closest integer before evaluating the objective. 

\begin{figure}
        \centering
        \includegraphics[width=.8\textwidth]{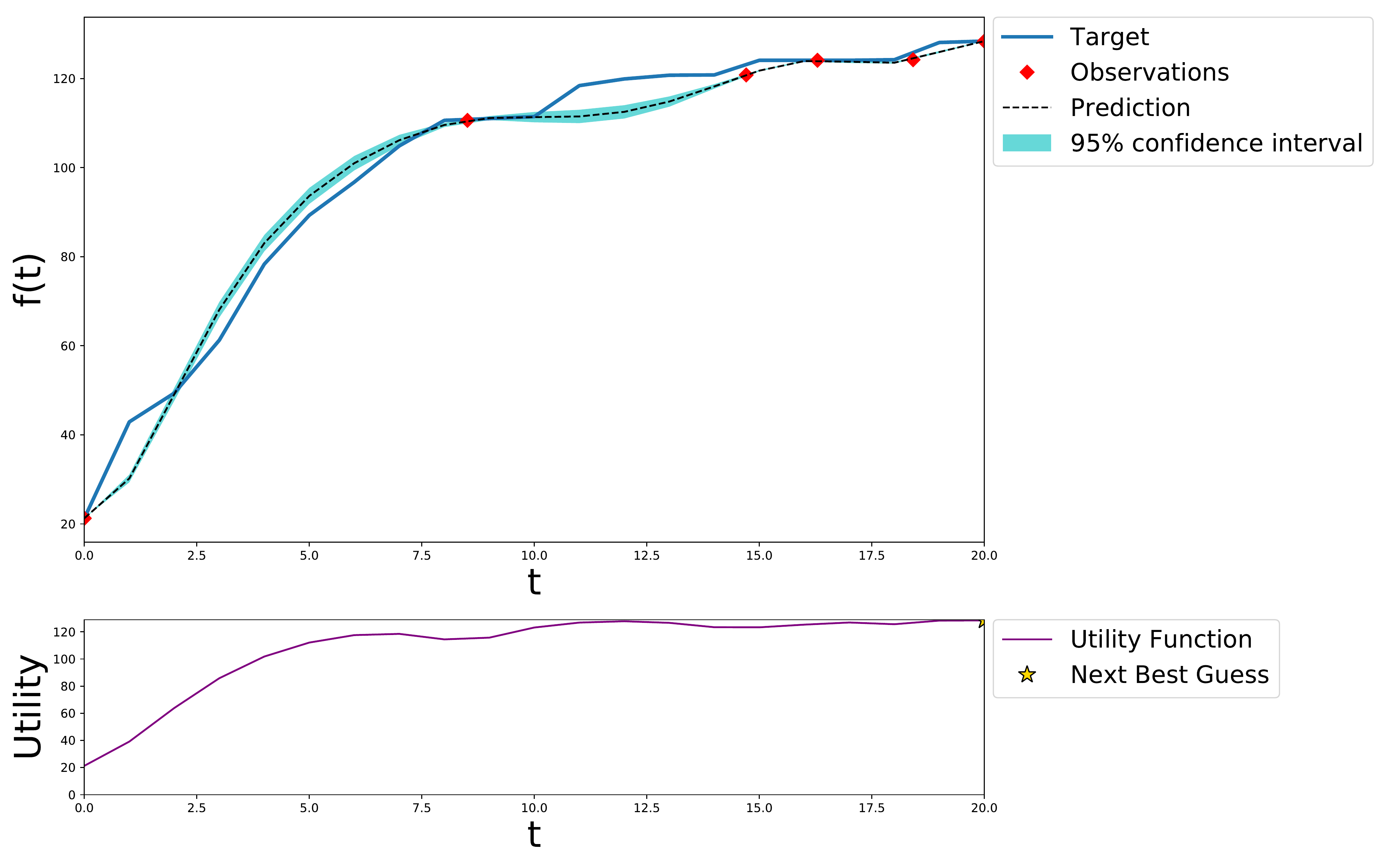}
        \caption{Above: Approximation of the objective function (\emph{target}) of one variable $t$ and fixed discrete parameters (${g: 2, k: 5}$), found after 3 random guesses and 3 iterations (\emph{observations}). The function is increasing for $t \in (0, 20]$ so the optimisation problem in this range is trivial. Below: The corresponding value of the utility function (Upper Confidence Bound, see Subsection \ref{GPUCB}) and the current estimation of the global maximum.}
\label{fig:oneparam}
\end{figure}

Instead of using approximations in the first set of experiments, we combined the exhaustive grid search approach for the two discrete parameters $g$, $k$ with Bayesian optimisation of the continuous parameter $t$. In the objective function we observed a plateau for $t > 20$ and so we optimised the parameter in the $(0, 20]$ range (see Figure \ref{fig:oneparam}).

We compared this approach with $\epsilon$-greedy search on the \textit{Polynomials} article of 135 conjectures and with grid search on all of the three parameters (with $t$ rounded to the nearest integer value). In this setting, the approach of employing Gaussian Processes gave the same results in terms of time and accuracy as $\epsilon$-greedy search. This is due to the simplicity of the objective function given one variable $t$ and the additional cost associated with the exhaustive search on parameters $g$ and $k$. Both probabilistic approaches reached the global maximum faster than a grid search using all three parameters. 

\begin{table}
\centering
\ra{1.3}
\begin{tabular}{@{}ccccc@{}}\toprule
AFP article & Nr of goals  & Proofs found [\%] & Time [s] & Optimal parameters  \\
 \midrule
 Polynomials & 135 & 87\% & 57s & {t: 16.3, g: 58, k: 131}\\ 
 AbstractHoareLogics & 793 & 63\% & 249s & {t: 17.6, g: 57, k: 130}\\ 
 Completeness & 475 & 89\% & 151s & {t: 18.9, g: 63, k: 134} \\
 FinFun & 263 & 95\% & 73s & {t: 19.6, g: 57, k: 132} \\
 HeardOf & 716 & 93\% & 331s & {t: 19.5, g: 57, k: 131} \\
 InductiveConfidentiality & 1425 & 82\% & 451s & {t: 19.6, g: 58, k: 130} \\
 RefineMonadic & 1509 & 95\% & 522s & {t: 14.7, g: 64, k: 123} \\
 MiniML & 345 & 84\% & 104s & {t: 19.1, g: 58, k: 131} \\
 RecursionTheory & 656 & 85\% & 205s & {t: 19, g: 57, k: 130} \\
 SortEncodings & 776 & 80\% & 437s & {t: 14, g: 64, k: 123} \\
 \bottomrule
\end{tabular}
\vspace{0.5pt}
\caption{Premise selection results on the AFP articles. For each article we performed three iterations of the Bayesian optimisation process with two random starting points. We report the time needed to complete this fixed number of iterations -- in several articles the optimum was found before the final iteration. }
\label{results}
\end{table}

\subsubsection{Bayesian Optimisation on the full set of parameters}
In the next set of experiments, we used the standard approach of approximating the discrete parameters with continuous variables, and optimised the objective function on the full set of parameters (see Table \ref{results}).

The $\epsilon$-search falls behind in this setting: we tested it for the three parameters within the time limits reported above, and the recommended parameters were in a close proximity to the initial random point, which yielded poor final results. We estimate that an exhaustive grid search in the comparable parameter space and on the same machine would take more than a day to find the global maximum.

\section{Conclusion}

The framework based on Bayesian optimisation with Guassian Processes turned out to be particularly effective in the multi-parameter setting. The main future goal is to evaluate the framework using Vampire and another mathematical corpus, for example TPTP \citep{Sutcliffe:2017:TPL:3158456.3158479}.

\bibliography{mybib}
\bibliographystyle{icml2018}

\end{document}